\documentclass{article}

    \PassOptionsToPackage{numbers, compress}{natbib}

\usepackage[preprint]{neurips_2024}

\usepackage[utf8]{inputenc} 
\usepackage[T1]{fontenc}    
\usepackage{hyperref}       
\usepackage{url}            
\usepackage{booktabs}       
\usepackage{amsfonts}       
\usepackage{nicefrac}       
\usepackage{microtype}      
\usepackage{xcolor}         
\usepackage{enumitem}
\usepackage{graphicx} 
\usepackage{amsmath} 
\usepackage{multicol}
\usepackage{amsmath,amssymb,amsfonts}
\usepackage{algorithmic}
\usepackage{graphicx}
\usepackage{textcomp}
\usepackage{xcolor}
\usepackage{mathtools}
\usepackage{amsmath}
\usepackage{array}
\usepackage{caption}
\usepackage{gensymb}
\usepackage{float}
\usepackage{adjustbox}
\usepackage{graphics}
\usepackage{multirow}
\usepackage{capt-of}
\usepackage{enumitem}
\usepackage{url}

\title{A Scientific Machine Learning Approach for Predicting and Forecasting Battery Degradation in Electric Vehicles}

\author{%
  Sharv Murgai\\
  Monta Vista High School\\
  Cupertino, CA \\
  \texttt{murgai.sharv@gmail.com} \\
    \And
  Hrishikesh Bhagwat \\
  Indian Institute of Technology Madras \\
  \texttt{bhagwatcoder@gmail.com} \\
   \AND
   Raj Abhijit Dandekar \\
  Vizuara AI Labs \\
  \texttt{raj@vizuara.com} \\
   \And
  Rajat Dandekar \\
   Vizuara AI Labs \\
  \texttt{rajatdandekar@vizuara.com} \\
  \And
  Sreedath Panat \\
   Vizuara AI Labs \\
  \texttt{sreedath@vizuara.com} \\
}

\begin{document}

\maketitle

\begin{abstract}

Carbon emissions are rising at an alarming rate, posing a significant threat to global efforts to mitigate climate change. Electric vehicles (EVs) have emerged as a promising solution to reduce emissions in the transportation sector, but their reliance on lithium-ion batteries introduces the critical challenge of battery degradation. Accurate prediction and forecasting of battery degradation over both short and long time spans are essential for optimizing performance, extending battery life, and ensuring effective long-term energy management. This directly influences the reliability, safety, and sustainability of EVs, supporting their widespread adoption and aligning with key United Nations Sustainable Development Goals (SDGs). In this paper, we present a novel approach to the prediction and long-term forecasting of battery degradation using a Scientific Machine Learning (SciML) framework that leverages Neural Ordinary Differential Equations (NeuralODEs) and Universal Differential Equations (UDEs). Unlike traditional machine learning models, which often function as black-box systems, SciML integrates domain knowledge with neural networks, offering more interpretable and scientifically grounded solutions for both predicting short-term battery health and forecasting degradation over extended periods. This hybrid approach captures both known and unknown degradation dynamics, improving predictive accuracy while reducing data requirements. We incorporate ground-truth data to inform our models, ensuring that both the predictions and forecasts reflect practical conditions. The model achieved a mean squared error of 9.90 with the UDE and 11.55 with the NeuralODE, in experimental data, a loss of 1.6986 with the UDE, and a mean squared error of 2.49 in the NeuralODE, demonstrating the enhanced precision of our approach. This integration of data-driven insights with SciML’s strengths in interpretability and scalability allows for robust battery management. By enhancing battery longevity and minimizing waste, our approach contributes to the sustainability of energy systems and accelerates the global transition toward cleaner, more responsible energy solutions, aligning with the UN’s SDG agenda.

\end{abstract}

\section{Introduction}

\subsection{Background and Motivation}

As of 2023, global carbon emissions have surged from 25.5 billion tonnes in 2000 to 37.55 billion tonnes \cite{b1}. Electric vehicles (EVs), which made up 18\% of total vehicle sales in 2023 \cite{b2}, are central to reducing emissions, representing a market worth 
\$500.48 billion \cite{b3}. Lithium-ion batteries power these EVs but face degradation over time, leading to reduced capacity and efficiency. This decline affects EV reliability and raises sustainability concerns. Addressing battery degradation is crucial for improving EV performance, extending battery lifespan, and supporting the broader adoption of sustainable transport solutions.

\subsection{Problem Statement}

Battery degradation refers to the irreversible decline in a battery's ability to store or deliver energy, characterized by reduced charge capacity and increased internal resistance. This is primarily driven by two forms of degradation: calendar degradation, which occurs over time, and cycle degradation, caused by repeated charging and discharging. Together, they impact the battery’s long-term performance. The state of health (SoH) quantifies this degradation as a percentage comparing current capacity to the original capacity when new. Monitoring SoH is crucial for extending battery life, especially in electric vehicles (EVs), allowing users to manage batteries more effectively, reduce replacements, conserve materials like lithium and cobalt, and support recycling and second-life applications.

However, predicting SoH remains challenging. Current models depend on empirical constants, which vary by battery chemistry, usage, and conditions, making universal models difficult. Validating these models requires long-term testing, often taking years. Additionally, the scarcity of long-term experimental data further complicates predictions, as many datasets are incomplete or short-term. To address this, we are utilizing high-quality experimental data from Sandia National Laboratories (SNL) to validate and enhance our model, aiming for more accurate and reliable SoH forecasting across various battery types.

\subsection{Contributions}
To accurately predict the SoH of EV batteries, we utilize a SciML \cite{7} framework that integrates physical models with deep learning techniques, offering more transparency than traditional black-box methods. In this paper, we introduce a UDE \cite{13} and a NeuralODE \cite{14} framework built to model and forecast battery SoH with upmost precision. The key objectives of this research are:
\begin{itemize}[leftmargin=0pt]
    \item Develop a UDE and NeuralODE framework that integrates ordinary differential equations (ODEs) with neural networks to model and predict battery health over time. \item Use synthetic ground-truth data and real-world experimental data from Sandia National Laboratories (SNL) \cite{6} to forecast battery degradation. \item Demonstrate the practical application of UDEs and NeuralODEs for long-term battery performance, focusing on efficiency and sustainability. \item Show how this approach supports global sustainability goals, aligning with the United Nations Sustainable Development Goals (SDGs) such as SDG 7, SDG 11, SDG 12, and SDG 13.
\end{itemize}

\section{Literature Review}

Wang et al. \cite{19} introduce a physics-informed neural network (PINN) to estimate the state-of-health (SoH) of lithium-ion batteries by combining empirical models with experimental data. While effective for cycle degradation, it overlooks calendar degradation and lacks Neural ODEs or UDEs, limiting its ability to generalize across battery chemistries. Hofmann et al. \cite{20} also develop a PINN, using data from labs, simulations, and vehicles to predict SoH, but similarly focus on cycle degradation and fail to incorporate physical laws with Neural ODEs or UDEs.

Ye et al. \cite{21} propose a Physics-Informed Neural Network (PIFNN) that integrates battery properties to improve SoH accuracy, but its reliance on specific datasets reduces generalizability. The absence of Neural ODEs or UDEs weakens its ability to handle complex degradation processes, particularly with noisy data. Anonymous \cite{4} employ a Scientific Machine Learning (SciML) framework with Universal Differential Equations (UDEs) to predict battery degradation, but lack of real-world validation remains a limitation.

\section{System Model and Methodology}

\subsection{The chemistry behind battery degradation}

To model lithium-ion battery degradation effectively, it’s important to understand the chemical processes involved. As shown in Figure 1, lithium-ion batteries consist of copper or aluminum current collectors, a graphite anode, and a 2,2,5,5-tetramethyloxolane (TMO) cathode. The two main forms of degradation are calendar degradation, which occurs over time, and cycle degradation, caused by repeated charge-discharge cycles. Calendar degradation involves the growth of the solid electrolyte interphase (SEI) layer on the anode, consuming lithium ions and reducing capacity, even when the battery is not in use. Cycle degradation occurs during charging and discharging, as lithium ions move between the anode and cathode, damaging the SEI layer and causing mechanical stress, cracks, and structural damage. This increases internal resistance and reduces capacity. High temperatures and voltage fluctuations accelerate these processes, while external factors like graphite exfoliation, SEI decomposition, and binder degradation further contribute to battery deterioration.

\begin{figure}
    \centering
    \includegraphics[width=0.5\linewidth]{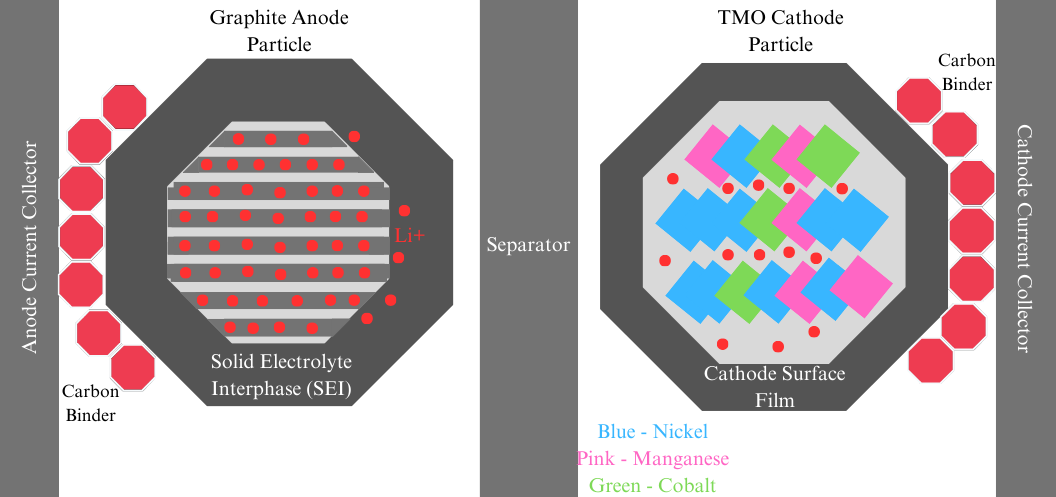}
    \caption{A schematic diagram showing the internal structure of a lithium ion battery}
    \label{fig:enter-label}
\end{figure}

\subsection{Modeling Battery Degradation: ODEs}
In this research, we have used the typical lithium-ion Nickel-Manganese-Cobalt (NMC) battery that is used in EVs, particularly a Nissan Leaf to accurately predict and forecast SoH.

\subsubsection{Calendar Degradation}
Calendar degradation occurs over time, even when the battery is not in use. It depends on the state of charge (SoC) \cite{24} of a battery, which represents the ratio of the current available charge (or energy) within a battery to its total charge capacity when fully charged, the time in days ($t$), the activation energy ($E_a$), which is the minimum amount of energy required to initiate a chemical reaction by enabling reactants to overcome the energy barrier and transition into products. Additionally, it is dependent on the internal temperature of the battery. The ODE governing calendar degradation is an Arrhenius formulation that quantifies the impact of temperature. In addition to these parameters, the equation uses a pre-exponential factor $f$, that is a piece-wise function of SoC as described in (1), and the values are outlined in Table 1. In this paper, we have used a base calendar degradation model from the Wang model \cite{11}. 

\begin{align}
    f(SoC)= \begin{dcases}
        -1.04 \cdot SoC^2(t) + 89.72 \cdot SoC(t) + 1224.6 \text{ if } SoC(t) \leq 50 \\
        10.35 \cdot SoC^2(t) - 1083.6 \cdot SoC(t) + 31447 \text{ if } 50 \leq SoC(t) \leq 70 \\
        2.64 \cdot SoC^2(t) - 409.55 \cdot SoC(t) + 22035 \text{ if } 70 \leq SoC(t) \\
    \end{dcases} \\
q_{cal}(t) = \int_{0}^{t} \frac{f(SoC)}{2} \cdot \frac{1}{\sqrt{t}} \cdot \exp^{\frac{-E_a}{R \cdot T_b}} \,dt
\end{align}

\begin{table}
\centering
\begin{tabular}{|c|c|c|c|c|c|c|}
\hline
\textbf{SoC (\%)} & 0 & 10 & 20 & 30 & 40 & 50 \\
\hline
\textbf{f} & 1224.6 & 2017.8 & 2603.0 & 2980.2 & 3149.4 & 3110.6 \\
\hline
\textbf{SoC (\%)} & 60 & 70 & 80 & 90 & 100 & - \\
\hline
\textbf{f} & 3691.0 & 6310.0 & 6167.0 & 6559.5 & 7480.0 & - \\
\hline
\end{tabular}
\caption{Values Corresponding to SoC (\%) and the pre-exponential factor $f$}
\label{table:soc_f_values}
\end{table}

Using the values of the pre-exponential function in Table 1, we can get the final ODE modeling calendar degradation as demonstrated in (2). 

where each of the parameters and variables are described in Table 2.

\begin{table}
\centering
\begin{tabular}{|c|c|c|}
\hline
\textbf{Parameter} & \textbf{Description} & \textbf{Value} \\
\hline
$f(\text{SoC})$ & Pre-exponential factor & Describes in equation 1 \\
$t$ & Time (days) & Battery life so far (duration of usage)\\
$E_a$ & Activation energy & 24500.0 $\text{J}\cdot\text{mol}^{-1}$ \\
$T_b$ & Battery temperature & 25$^\degree$C ($298.0$K) \\
$R$ & Gas constant & 8.314 $\text{J}\cdot\text{mol}^{-1}\cdot\text{K}^{-1}$ \\
\hline
\end{tabular}
\caption{Parameters and descriptions for calendar degradation}
\label{table:parameters}
\end{table}
\subsubsection{Cycle Degradation}
Cycle degradation occurs over time due to charge and discharge cycles, and is dependent on discharge current ($I_b$), battery temperature ($T_b$), and the battery capacity ($Q$). In this paper, we have used the Wang model \cite{11} to accurately model the cycle degradation of the specific NMC lithium-ion battery used in Nissan Leafs.  The parameters and variables are described in Table 3. In order to calculate the discharge current of the battery, we can use (4) with the parameters as shown in Table 4 \cite{10}.

\begin{align}
    q_{cycl} = \int \left( a \cdot T_b^2 + b \cdot T_b + c \right) \cdot \exp^{\left( \frac{d \cdot T_b + e}{Q} \cdot I_b \right)} \cdot \frac{I_b}{Q \cdot 3600} \, dt
    \\
    I_b = \frac{1}{V_{\text{nom}}} \cdot \frac{\text{Odo} \cdot \eta}{\Delta t_{\text{driving}}}
\end{align}


\begin{table}
\centering
\begin{tabular}{|c|c|c|}
\hline
\textbf{Parameter} & \textbf{Description} & \textbf{Value} \\
\hline
$a$ & Empirical coefficient in $\text{1/Ah}\text{K}^2$ & $8.61 \times 10^{-6}$ \\
$b$ & Empirical coefficient in $\text{1/Ah}\text{K}$ & $-5.13 \times 10^{-3}$ \\
$c$ & Empirical coefficient in $\text{1/Ah}$ & $7.63 \times 10^{-1}$ \\
$d$ & Empirical coefficient in $\text{1/K}\text{(C-rate)}$ & $-6.7 \times 10^{-3}$ \\
$T_b$ & Battery temperature & 25$^\degree$C ($298.0$K) \\
$e$ & Empirical coefficient in $\text{1/(C-rate)}$ & $2.35$ \\
$I_b$ & Discharge current in Amperes  & Eq 4\\
$Q$ & Battery capacity in Amp-Hours (Ah)  &  176.4 Ah\\
\hline
\end{tabular}
\caption{Parameters and their descriptions for cycle degradation}
\label{table:parameters}
\end{table}

In the absence of real-world testing conditions and to simplify the model in the early stages of our research, we made several assumptions regarding average driving time and mileage. These assumptions were based on data from multiple studies conducted in China \cite{26}, which indicate that the typical annual driving distance is approximately 20,000 kilometers. Using (3) and (4), we can calculate cycle degradation. 

\begin{table}
\centering
\begin{tabular}{|c|c|c|}
\hline
\textbf{Parameter} & \textbf{Description} & \textbf{Value} \\
\hline
$V_\text{nom}$ & Battery nominal voltage & 350.4 V \\
$\text{Odo}$ & Odometer readings & Average of 60kms/day\\
$\eta$ & Driving specific energy consumed & 180 Wh/km \\
$\Delta t_{\text{driving}}$ & Driving time & Average of 2hrs \\
\hline
\end{tabular}
\caption{Parameters and their descriptions for discharge current}
\label{table:parameters}
\end{table}

\subsubsection{Cumulative Degradation}
Based on (1), (2), (3) and (4), we can calculate the cumulative degradation, capacity and the SoH of the battery using (5), (6) and (7):
\begin{align}
    {q_{total}} &= q_{cycl} + q_{cal} \\
    SoH &= 100\% - q_{total} \\
    Q &= Q_{nom} \cdot (100\% - q_{total}),
\end{align}
    
where $Q_{nom}$ is the maximum capacity of the battery when it was not used when $t = 0$, around 176.4 Ah. 

\begin{figure}[H]
\centering
\includegraphics[width=1.0\linewidth]{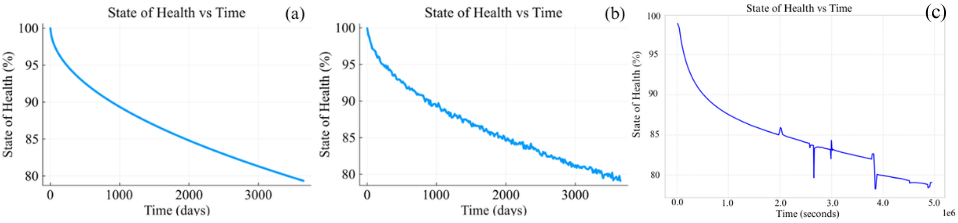}
\captionof{figure}{(a) Simulation of SoH over a timespan of 10 years; (b) and with gaussian noise; (c) Experimental data}
\end{figure}

After running simulations under these conditions using the DifferentialEquations.jl \cite{27} package in Julia, as outlined in previous sections, the results indicate a total degradation of approximately 20\%, as illustrated in Figure 2(a), which aligns with real-world expectations. At an optimal temperature of 25°C, cycle degradation contributes only about 1.5\% to the overall degradation, with the majority attributed to calendar effects.

\subsection{Prediction of SoH: synthetic ground-truth data}
In the initial stage of the project, we lacked experimental ground-truth data, so we generated synthetic data using the ground-truth equations (2), (3), (4), and (5). Gaussian noise with an amplitude of 0.20 was added to the cumulative degradation (Figure 2(a)) to simulate real-world conditions, resulting in the data shown in Figure 2(b). This noisy dataset serves as a benchmark for predictions made by the UDE and NeuralODE models. We now explore SciML frameworks, including UDEs and NeuralODEs, to forecast battery degradation over time. 

\begin{itemize}[leftmargin=0pt]
    \item \textbf{UDE}: A UDE combines traditional mathematical models with neural networks, blending physics-based knowledge with data-driven methods. This hybrid approach allows UDEs to capture complex system dynamics, effectively modeling systems with both known physical laws and data-dependent behaviors.

    For the UDE, we chose to replace the $\frac{1}{\sqrt{t}}$ term from (2), and the $\left( a \cdot T_b^2 + b \cdot T_b + c \right) \cdot \exp^{\left( \frac{d \cdot T_b + e}{Q} \cdot I_b \right)} \cdot \frac{I_b}{Q \cdot 3600}$, from (3), or the cycle degradation expression with neural networks. The resultant equation is described in (8), where $\text{NN}_1(t)$ is a neural network that approximates $\frac{1}{\sqrt{t}}$ and $\text{NN}_2(T_b,I_b,Q)$ is a neural network that approximates the $\left( a \cdot T_b^2 + b \cdot T_b + c \right) \cdot \exp^{\left( \frac{d \cdot T_b + e}{Q} \cdot I_b \right)} \cdot \frac{I_b}{Q \cdot 3600}$ term. 
    
    \begin{equation}
        \frac{dq_{total}}{dt} = \frac{f(SoC)}{2} \cdot \text{NN}_1(t) \cdot \exp^{\left(\frac{-E_a}{R \cdot T_b}\right)} + \text{NN}_2(T_b, I_b, Q)
    \end{equation}

    The $\frac{1}{\sqrt{t}}$ term was replaced with a neural network to capture the time dependency of calendar degradation more accurately, which is crucial for understanding long-term battery performance. This allows the UDE to learn the time-dependent behavior, enhancing the SciML framework. Similarly, the cycle degradation term from equation (3) was replaced with a neural network, reducing reliance on empirical constants and enabling the model to learn complex patterns directly from data, making the degradation process more flexible and precise. Training the UDE on synthetic ground-truth data was done in two stages. Each neural network had one input neuron and three hidden layers (10, 5, and 5 neurons) with ReLU activation. In the second stage, batch normalization was applied to each hidden layer for faster and more stable training, and a dropout layer (10\%) was used to reduce overfitting. These steps improved the overall performance of the UDE.
    
    \item \textbf{NeuralODE}: A NeuralODE uses neural networks to learn continuous-time dynamics by solving ODEs. Unlike discrete layers, it models changes over time, making it ideal for time-series prediction and system dynamics. The neural network defines the system's derivative, which is integrated over time by an ODE solver for flexible, efficient modeling.

    \begin{align}
        \frac{dq_{total}}{dt} &= f(\mathbf{h}(t), t, \theta)
    \end{align}
    where \( q_{total} \) represents the total state of the system, \( \frac{dq_{total}}{dt} \) is the time derivative of the system state, \( \mathbf{h}(t) \) is the hidden state of the system at time \( t \), and \( f(\mathbf{h}(t), t, \theta) \) is a neural network parameterized by \( \theta \), which captures the system dynamics. The NeuralODE for predicting battery degradation, implemented in Julia's Flux.jl \cite{28}, features an input layer, three hidden layers (32, 64, and 32 neurons with ReLU activation), and an output layer. Key inputs include time, temperature, SoC, current, capacity, voltage, and cycle count. Batch normalization ensures stability, while dropout (10-20\%) mitigates overfitting. The output layer predicts the time derivative of total degradation $\frac{dq_{total}}{dt}$, which is integrated using an ODE solver to compute cumulative degradation. This architecture, combined with DiffEqFlux.jl \cite{29}, effectively models battery degradation dynamics.

\end{itemize}

\subsection{Prediction of SoH: Using real-world experimental data}

\subsubsection{Dataset}

The dataset from Sandia National Labs \cite{5} is a key resource for studying lithium-ion NMC battery degradation. It includes experimental data on cycle count, capacity, voltage, current, and temperatures. NMC cells (3Ah capacity) were cycled to end of life (80\% SOH) at various temperatures (15 °C, 25 °C, 35 °C), depths of discharge (0–100\%, 20–80\%, 40–60\%), and discharge rates (0.5C, 1C, 2C, 3C). After basic preprocessing to remove outliers and missing values, we generated a plot (Figure 2(c)) showing capacity degradation over time under these conditions.


\section{Experiments and Results}
\subsection{Training}
For experiments on both the NeuralODE and UDE, hyperparameter tuning was performed, testing various timespans, optimizers, and learning rates. Optimizers like Adam \cite{30}, Nesterov \cite{31}, AdaBelief \cite{32}, AdaGrad \cite{33}, Sophia \cite{34}, and RMSProp \cite{35} were evaluated with learning rates of 0.1, 0.01, 0.001, and 0.0001, over 10,000 to 80,000 iterations, in increments of 10,000. The UDE and NeuralODE was trained on timespans from 1 to 10 years, and the MSE was recorded for each setup to determine the best configuration. To evaluate the preformances of the models, MSE was used.

\subsubsection{Synthetic ground-truth data: UDE}
\begin{table}[H]
\centering
\begin{tabular}{|c|c|c|c|c|}
\hline
\textbf{TimeSpan* (days)} & \textbf{Optimizer} & \textbf{StepSize} & \textbf{Iterations} & \textbf{Loss} \\
\hline
(1.0, 365.0) & RMSProp  & 0.0001 & 10000 & 12.321524 \\
(1.0, 730.0) & ADAM     & 0.1 & 50000 & 27.327161 \\
(1.0, 1095.0) & Sophia  & 0.01 & 80000 & 46.017585 \\
(1.0, 1460.0) & AdaBelief & 0.01 & 50000 & 65.239215 \\
(1.0, 1825.0) & Nesterov & 0.01 & 70000 & 81.351090 \\
(1.0, 2190.0) & ADAM     & 0.01 & 80000 & 101.366405 \\
(1.0, 2555.0) & ADAM     & 0.01 & 30000 & 112.077484 \\
(1.0, 2920.0) & AdaGrad  & 0.01 & 70000 & 134.881377 \\
(1.0, 3285.0) & Sophia   & 0.001 & 50000 & 151.313181 \\
(1.0, 3650.0) & Nesterov & 0.001& 70000 & 174.597625 \\
\hline
\end{tabular}
\caption{Optimization Results for UDE training (Experiment 1). *TimeSpan is from day 1 to specified end day}
\label{table:ude_training_results}
\end{table}

\begin{table}
\centering
\begin{tabular}{|c|c|c|c|c|}
\hline
\textbf{TimeSpan (days)} & \textbf{Optimizer} & \textbf{StepSize} & \textbf{Iterations} & \textbf{Loss} \\
\hline
(1.0, 365.0)  & Nesterov  & 0.01 & 80000 & 4.172145  \\
(1.0, 730.0)  & RMSProp   & 0.01 & 80000 & 4.874884  \\
(1.0, 1095.0) & ADAM      & 0.01 & 60000 & 6.640595  \\
(1.0, 1460.0) & ADAM      & 0.01 & 50000 & 55.925221 \\
(1.0, 1825.0) & Sophia    & 0.01 & 80000 & 27.453747 \\
\textbf{(1.0, 2190.0)} & \textbf{AdaBelief} & \textbf{0.01} & \textbf{60000} & \textbf{9.901479}  \\
(1.0, 2555.0) & ADAM      & 0.01 & 60000 & 24.132372 \\
(1.0, 2920.0) & AdaGrad   & 0.01 & 40000 & 27.752057 \\
(1.0, 3285.0) & ADAM      & 0.01 & 70000 & 38.811691 \\
(1.0, 3650.0) & ADAM      & 0.01 & 50000 & 32.714843 \\
\hline
\end{tabular}
\caption{Optimization Results for UDE training with batch normalization and dropout layers (Experiment 2)}
\label{tab:optimization_results}
\label{table:ude_training_results_2}
\end{table}
\vspace{-20pt}


\begin{figure}
    \centering
    \includegraphics[width=0.8\linewidth]{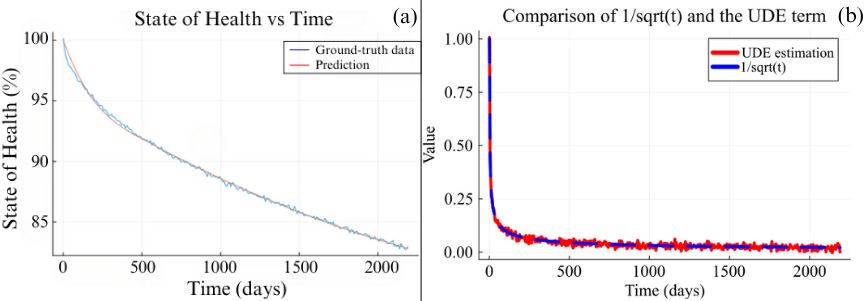}
    \caption{(a) UDE prediction over a 6-year timespan using ADAM; (b) Comparison of $\frac{1}{\sqrt{t}}$ term to the UDE term}
    \label{fig:enter-label}
\end{figure}

In the first experiment (Table 5), without dropout or batch normalization, high losses occurred for timespans over 5 years and lacked accuracy for shorter spans. A second experiment introduced batch normalization and dropout, reducing overfitting and improving MSE across all timespans (Table 6). The UDE achieved an MSE of 9.90 over 6 years, closely matching ground-truth data (Figure 3(a)). Finally, we compared the UDE estimation with the 
term (Figure 3(b)), confirming that the UDE accurately captures the time-dependent degradation of the battery.

\subsubsection{Synthetic ground-truth data: NeuralODE}
In order to evaluate the performance of the NeuralODE, we also use MSE across all the iterations. Only one experiment was conducted, which gave a MSE of 11.55 over a 7-year timespan, as demonstrated in Figure 4. The results of the experiment are shown in Table 7. Batch normalization and dropout layers were included from the start. There were lower losses over smaller timepsans. However, they are not enough to perform adequate forecasting, and nondeterministic to make conclusions. 

\begin{figure}
    \centering
    \includegraphics[width=0.35\linewidth]{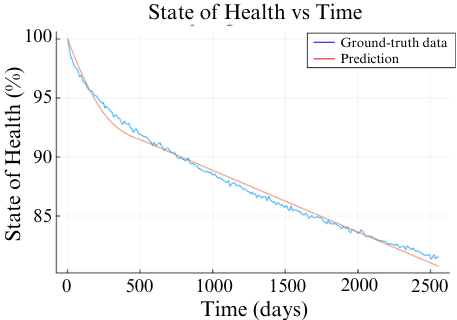}
    \caption{NeuralODE prediction over 7-year timespan using ADAM.}
    \label{fig:enter-label}
\end{figure}

\begin{table}
\centering
\begin{tabular}{|c|c|c|c|c|}
\hline
\textbf{TimeSpan (days)} & \textbf{Optimizer} & \textbf{StepSize} & \textbf{Iterations} & \textbf{Loss} \\
\hline
(1.0, 365.0)  & Nesterov      & 0.01 & 80000 & 5.23  \\
(1.0, 730.0)  & AdaBelief   & 0.001 & 70000 & 8.79  \\
(1.0, 1095.0) & Adam  & 0.001 & 60000 & 6.51  \\
(1.0, 1460.0) & RMSProp      & 0.01 & 50000 & 12.77 \\
(1.0, 1825.0) & Sophia    & 0.01 & 80000 & 19.87 \\
(1.0, 2190.0) & RMSProp   & 0.001 & 60000 & 22.32 \\
(1.0, 2555.0) & Adam  & 0.001 & 70000 & 27.98 \\
\textbf{(1.0, 2555.0)} & \textbf{AdaBelief} & \textbf{0.01} & \textbf{60000} & \textbf{11.55} \\
(1.0, 2920.0) & AdaGrad   & 0.01 & 40000 & 30.12 \\
(1.0, 3285.0) & RMSProp      & 0.001 & 80000 & 35.78 \\
(1.0, 3650.0) & ADAM      & 0.01 & 50000 & 27.71 \\
\hline
\end{tabular}
\caption{Optimization Results for NeuralODE training on synthetic data with batch normalization and dropout layers}
\label{tab:optimization_results}
\end{table}
\vspace{-10pt}

\subsubsection{Experimental data: UDE}

Using the pre-processed dataset, we performed an 80-20 train-test split, with the training set comprising 45,769 records. Adam provided the best performance with a loss of 1.6980, while other optimizer results are shown in Table 8. The UDE model accurately captured time-dependent and cycle degradation dynamics, as shown in Figure 5(a). This prediction improves battery health monitoring, enabling earlier decisions on repurposing, recycling, or extending battery life, reducing the need for frequent replacements. By minimizing resource extraction and waste, it supports global sustainability efforts, aligning with the UN’s 9th, 11th, 12th, and 13th SDGs.

\begin{figure}
    \centering
    \includegraphics[width=0.94\linewidth]{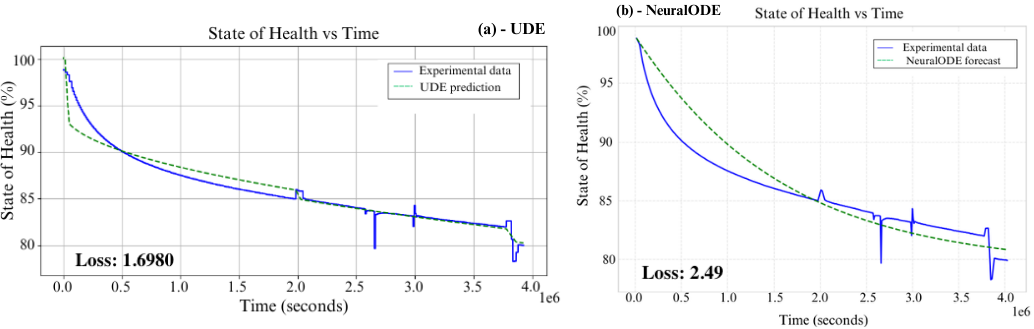}
    \caption{(a) UDE prediction over experimental data using ADAM; (b) NeuralODE prediction over experimental data using ADAM and DOPRI5 solver.}
    \label{fig:enter-label}
\end{figure}


\begin{table}
\begin{multicols}{2} 
    \centering
    \begin{tabular}{|c|c|}
    \hline
    \textbf{Optimizer (UDE)} & \textbf{Loss} \\
    \hline
    Nesterov   & 7.1243 \\
    RMSProp    & 23.6789 \\
    \textbf{Adam}       & \textbf{1.6980} \\
    AdaGrad    & 14.5231 \\
    AdaBelief  & 11.9874 \\
    Sophia     & 6.2456 \\
    \hline
    \end{tabular}
    \caption{Optimizer vs Loss Comparison for UDE training on Experimental Data}
    \label{tab:optimizer_loss_ude}

    \columnbreak 

    \centering
    \begin{tabular}{|c|c|}
    \hline
    \textbf{Optimizer (NeuralODE)} & \textbf{Loss} \\
    \hline
    Nesterov   & 20.10 \\
    RMSProp    & 14.219 \\
    \textbf{Adam}       & \textbf{2.1} \\
    AdaGrad    & 10.14 \\
    AdaBelief  & 9.10 \\
    Sophia     & 8.20 \\
    \hline
    \end{tabular}
    \caption{Optimizer vs Loss Comparison for NeuralODE training on Experimental Data}
    \label{tab:optimizer_loss_neuralode}
\end{multicols}
\end{table}

\subsubsection{Experimental data: NeuralODE}
Using the preprocessed dataset, we conducted an 80-20 train-test split to predict battery degradation over the final 20\% of the data. The training set comprised 45,769 records. After testing multiple optimizers and the DOPRI solver, Adam demonstrated the best performance with training MSE loss of 2.49 as shown in Figure 5(b), while the results of other optimizers are summarized in Table 9.


\subsection{Battery Degradation Forecasting}
The UDE, trained on synthetic data over 6 years, achieved an MSE of 9.90, while the NeuralODE reached 11.55 (Figure 3(a)). Forecasting over 10 years is critical for sustainable EV battery management. On experimental data, the UDE achieved an MSE of 1.6980, closely matching real-world SoH values (Figure 6(a)), while the NeuralODE had an MSE of 2.1 (Figure 7). These long-term predictions enhance battery life, support timely recycling, and reduce resource demand, aligning with the UN's 9th, 11th, 12th, and 13th sustainability goals. The UDE’s forecast of ~80\% SoH after 10 years (Figure 6(b)) highlights its capability for sustainable battery management.

\begin{figure}
    \centering
    \includegraphics[width=0.9\linewidth]{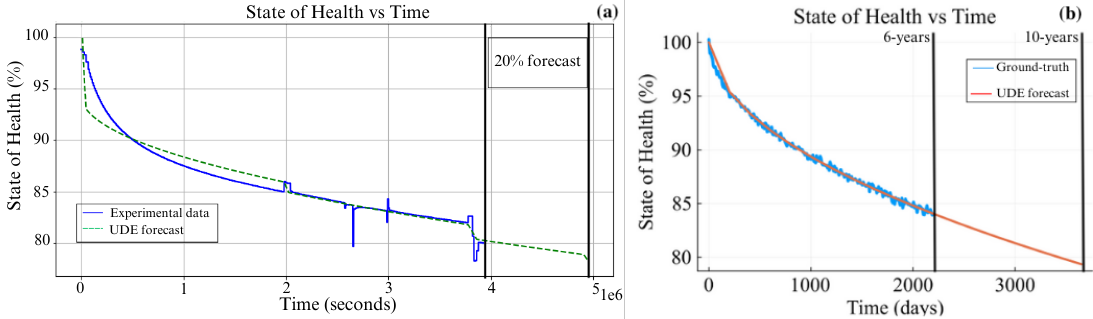}
    \caption{(a) UDE forecast on 20\% of experimental data over full timespan; (b) Forecasting of the UDE over 10 years on synthetic ground-truth data}
    \label{fig:enter-label}
\end{figure}

\begin{figure}
    \centering
    \includegraphics[width=0.45\linewidth,scale=0.5]{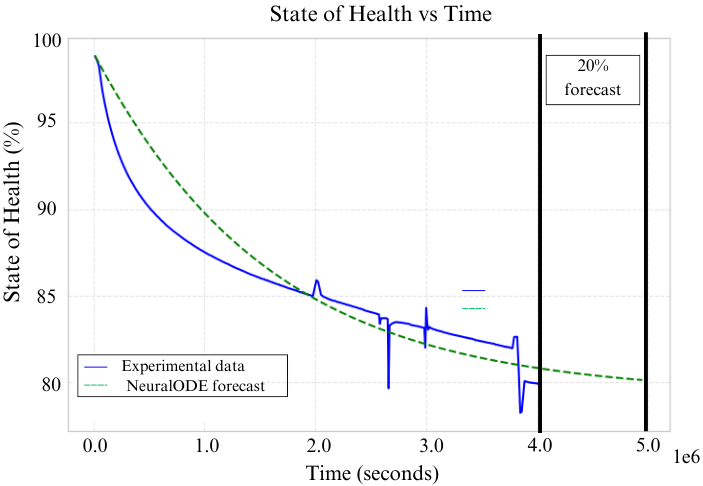}
    \caption{NeuralODE forecast on 20\% of experimental data over full timespan}
    \label{fig:enter-label}
\end{figure}
\section{Conclusion}

In this work, we explained the chemical processes behind calendar and cycle degradation, including SEI growth, particle cracking, and anode dissolution. We employed Scientific Machine Learning (SciML) to predict and forecast battery degradation using NeuralODE and UDE models, achieving MSE losses of 9.90 and 11.55 on synthetic data, and 1.6980 and 2.10 on real-world data, respectively. SciML integrates machine learning with physical models, offering more accurate and interpretable predictions by capturing complex battery dynamics. Unlike traditional black-box methods, which often overlook the physical processes driving battery degradation, SciML leverages domain knowledge to improve generalization, making it an essential tool for battery health forecasting and advancing sustainable energy storage solutions.

\medskip

\small

‌


\end{document}